\documentclass[12pt]{article}

\usepackage{scicite}

\usepackage{times}
\usepackage[T1]{fontenc} 
\usepackage{graphicx}
\usepackage{amsmath,amsfonts,amscd,amssymb}
\usepackage{verbatim}
\usepackage{hyperref}
\usepackage{multirow}

\newenvironment{sciabstract}{%
\begin{quote} \bf}
{\end{quote}}

\newcounter{lastnote}

\usepackage{xr}
\makeatletter

\newcommand*{\addFileDependency}[1]{
\typeout{(#1)}
\@addtofilelist{#1}
\IfFileExists{#1}{}{\typeout{No file #1.}}
}\makeatother

\newcommand*{\myexternaldocument}[1]{%
\externaldocument{#1}%
\addFileDependency{#1.tex}%
\addFileDependency{#1.aux}%
}

\myexternaldocument{supplement}

\title{Computed tomography using meta-optics}

\author
{Maksym Zhelyeznyakov,$^{1\ast}$ Johannes E. Fr{\"o}ch,$^{1}$, Shane Colburn$^1$, Steven L. Brunton$^3$,\\ Arka Majumdar$^{1\ast,2}$\\
\\
\normalsize{$^{1}$Department of Electrical and Computer Engineering, University of Washington,}\\
\normalsize{Seattle, Washington, 98195, USA}\\
\normalsize{$^{2}$Department of Physics, University of Washington,}\\
\normalsize{Seattle, Washington, 98195, USA}\\
\normalsize{$^3$ Department of Mechanical Engineering, University of Washington,}\\
\normalsize{Seattle, Washington, 98195, USA}
\\
\normalsize{$^\ast$ E-mail: mzhelyez@gmail.com}
}

\date{}

\begin{document} 


\baselineskip24pt


\maketitle


\begin{sciabstract}
Computer vision tasks require processing large amounts of data to perform image classification, segmentation, and feature extraction. Optical preprocessors can potentially reduce the number of floating point operations required by computer vision tasks, enabling low-power and low-latency operation. However, existing optical preprocessors are mostly learned and hence strongly depend on the training data, and thus lack universal applicability.  In this paper, we present a metaoptic imager, which implements the Radon transform obviating the need for training the optics. High quality image reconstruction with a large compression ratio of $0.6\%$ is presented through the use of the Simultaneous Algebraic Reconstruction Technique. Image classification with $90\%$ accuracy is presented on an experimentally measured Radon dataset through neural network trained on digitally transformed images.
\end{sciabstract}


\section*{Introduction}
Mathematical transforms such as Fourier, Laplace, and wavelet transforms play a fundamental role in signal and image processing, telecommunications, and machine learning. These transforms relate functions and their properties across different domains. Traditionally these transforms rely on digital computation, which involves a large number of steps, and thus incur significant power consumption and latency. Free space optical systems are inherently well suited to perform these mathematical operations due to their high spatial, bandwidth, and inherent parallelism. Various optical systems have been devised to perform mathematical transforms, such as the 4f system, which is well known to perform an optical analogue of the Fourier transform and convolution \cite{goodman}. Others include the cosine transform \cite{optical-cos-transform, optical-cos-transform-2},  Laplace transform \cite{opt-laplace},  wavelet transform \cite{opt-wavelet}, and Radon transform \cite{opt-radon}. All of these systems, however, require bulky optical components, thus imposing a large footprint.

Two-dimensional meta-optics, a subclass of diffractive optics, with sub-wavelength features, have shown great promise in both miniaturizing existing optical elements \cite{Zhan2016, Wang2018}, and developing optical elements with new functionalities \cite{Zhan2017, Colburn:18}. Furthermore, a combination of meta-optics and computational imaging has also created a new research direction over the past decade. Here, a meta-optical front end pre-processes a scene to capture an intermediate image, and then a computational back end either creates an aesthetically pleasing image or extracts information relevant for computer vision. Such synergistic hybrid meta-optical/digital systems, also termed as "software-defined meta-optics" \cite{SW-defined-mo} have primarily been employed for broadband\cite{Colburn-SciAdv,Tseng2021,froch2024,Dong2024,Maman2023} and multi-functional imaging/sensing\cite{froch2023,Colburn2020,YangLinChen,Guo2019-vl,Shen2023}. Recently, similar systems are being explored for performing computer vision applications, including object detection and classification\cite{Zhang2022,Huang2024,Zheng2024,Anna2024,Wei2023}. Here, the fundamental idea is to extract features from the scene, and to perform a part of the computation directly on the light from the scene. 

Most prior works that utilize meta-optics as an optical front end to computer vision systems, however, primarily focus on implementing convolution operations, and performing some of the initial linear operations of an artificial neural network using optics \cite{Fu2024,d2nn,Wirth-Singh:24}. While such learned convolutional operations have resulted in a drastic reduction in required digital operations (such as multiply-and-accumulate), the convolutional operations are very much dataset dependent and non-generalizable. Hence, when the dataset is changed, either a completely new set of convolutional kernels is needed, or the digital back end must be retrained. This poses a serious limitations, as reconfiguring a phase mask at will requires a significant amount of power, obviating the power and latency benefit of optical frontend. While a large body of works exist on reconfigurable meta-optics, very few actually demonstrated independent control of the meta-atoms, and even then they are mostly limited to 1D\cite{Fang2024-wn,Park2021-ea,Shirmanesh2020-mj}. The controlled 2D array is limited to few pixels\cite{Moitra2023-gg}, and scaling such reconfigurable sub-wavelength phase masks poses several fundamental challenges\cite{Chen2023-ut}.

One promising solution could be to employ optics, that extract the features from the scene, but do not depend on the image itself, i.e., not a learned front end. In fact, researchers have employed random optics, followed by a computational back end for image classification\cite{Bai2023,Gao2021,OSTROVSKY200343,Hanawa:22}. However, such an approach requires calibrating each random surface. In this paper, we implement the Radon transform using a meta-optics.  The Radon transform is implemented under incoherent light from an organic light emitting diode display using a meta-optic. We first demonstrate high-quality image reconstruction, which requires the capture of only $\sim 0.6\%$ of the pixels in the reconstructed image. We then use an artificial neural network to directly classify the image. We train the image using simulated result of the Radon transform, and achieved a high classification accuracy with experimentally measured Radon transformed data without retraining.

\begin{figure}[t]
\centering
\includegraphics[width=\textwidth]{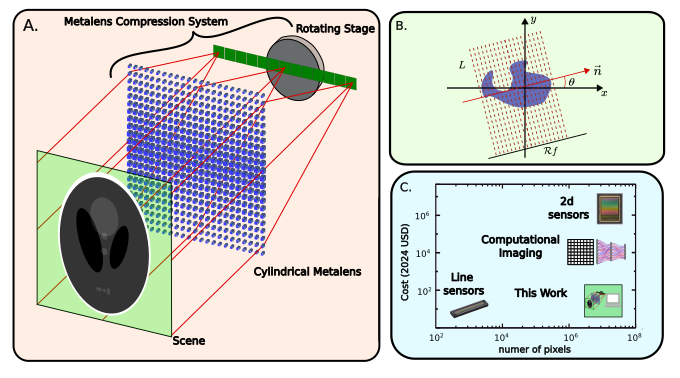}
\caption{\textbf{a.} Schematic for meta-optic computed tomography. An object is imaged with cylindrical metalens at the back-focal-plane along the focal line, measuring the DC intensity component of 1-D Fourier transforms along different angles corresponding to different slices in Radon space. \textbf{b.} Schematic of Radon transform operation. Integrals along straight lines $L \subset \mathbb{R}^2$ are taken at different rotation angles $\theta$ to construct $\mathcal{R}f$ \textbf{c.} Comparison of cost vs number of pixels of different imaging systems. Our system is low cost and can achieve imaging with a very high number of pixels.}
\label{fig:rt-schematic}
\end{figure}
\newpage
\section*{Methods}
The Radon transform was originally introduced by Johann Radon in 1917 \cite{OGRadon} for pure mathematical reasons when studying integral geometries. His work went largely unnoticed until 1972 when it was found to be important in radiological applications \cite{CT-Invention}. In two dimensions the Radon transform is defined on the set of rotated $L \subset \mathbb{R}^2$. The x and y coordinates of a straight line in $L$ can be parameterized a coordinate $z$ that lies on the line:
\begin{equation}
\begin{split}
    x(z) = z \sin(\theta) + s \cos(\theta) \\
    y(z) = - z \cos(\theta)  + s \sin(\theta)
\end{split}
\end{equation}
In the above equation $s$ is the distance of $L$ from the origin and theta is the angle the normal vector of $L$ makes with the x-axis. Thus, we can define a two-dimensional Radon transform of a function $f(x,y)$ by
\begin{equation}
    \tilde{f}(s, \theta) = \mathcal{R}f= \int_{-\infty}^\infty f(x(z), y(z)) dz
\end{equation}
A visual diagram of the Radon transform is shown in \autoref{fig:rt-schematic}B. This integral can be interpreted as taking the average value of the function along the line of integration. Serendipitously, when imaging a scene at the back focal plane of a cylindrical lens, this functionality is achieved when recording the pixel values along the focal line of the lens as shown in \autoref{fig:rt-schematic}A. Thus the Radon transform can be implemented optically by using a cylindrical lens and a line detector, and rotating the system to record the line-average of the scene at different angles. We note that due to the use of a line detector, our system provides a promising alternative to expensive, high pixel density 2D optical sensors, as shown in \autoref{fig:rt-schematic}C.

To optically implement the Radon transform, we designed a metalens with a cylindrical phase profile, operating at $780nm$ wavelength. The phase profile is given as:
\begin{equation}
    \phi_{lens} (x, y) = \frac{2\pi}{\lambda} ( F - \sqrt{F^2 + x^2} )
\label{eq:ml-phase}
\end{equation}
In the above equation, $x$ and $y$ are the spatial coordinates of the metalens and $F$ is the focal length of the metalens. Note that, since this is a cylindrical metalens, the phase profile only varies in one direction, chosen to be $x$ in our case. We used crystalline Silicon on Sapphire substrate as the material platform for the metalens. The scatterer geometry used in this work is square pillars with height $500nm$, at periodicity of $330nm$. We computed the amplitude and phase response of the scatterers using Rigorous Coupled Wave Analysis (RCWA). The constructed electromagnetic response vs scatterer geometry library is shown in \ref{fig:s1}. Due to the manufacturing constraints, we constrained the pillar diameters to vary between $70$ and $200nm$. To design the meta-optic, we first discretized the phase profile in equation \ref{eq:ml-phase} by the pillar periodicity on a 2 dimensional $x-y$ grid. At every point $(x_i, y_i)$, we matched the computed lens phase profile $\phi_{lens}(x_i,y_i)$ to the pillar with the closest phase response:
\begin{equation}
d(x_i,y_i) = \min_d || \phi_{RCWA}(d) - \phi_{lens}(x_i,y_i)||_2^2
\label{eq:ml-design}
\end{equation}

\begin{figure}[t]
\centering
\includegraphics[width=\textwidth]{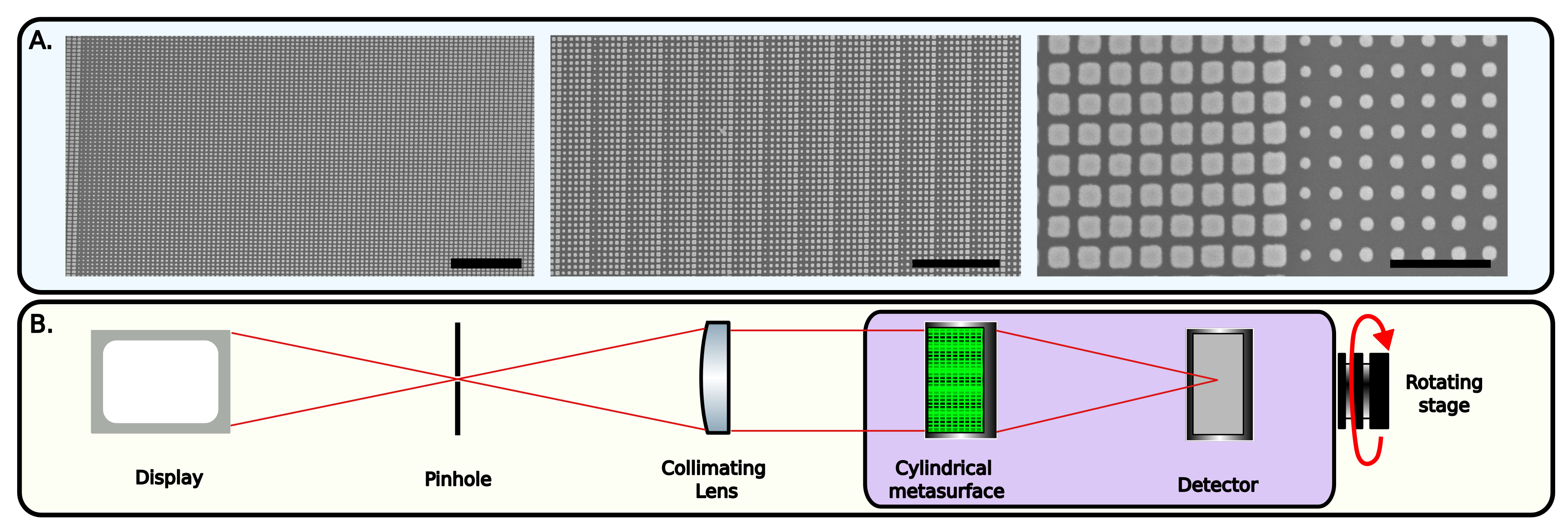}
\caption{\textbf{A.} Scanning electron micrographs (SEM) of fabricated metalens. Scale bars correspond to $5\mu m$ for the left most SEM image, $5\mu m$ for the center, and $1\mu m$ for the rightmost. \textbf{B.} Experimental setup. A display projects an image through a pinhole. A lens is used to collimate light from the pinhole onto the cylindrical metasurface. The Cylindrical metasurface and the detector are mounted onto a rotating stage, which rotates the system to measure the image at different angles.}
\label{fig:ms-design}
\end{figure}

The pillar distribution of the meta-optic is shown in \autoref{fig:s1}-. Scanning electron micrographs of the manufactured device are shown in \autoref{fig:ms-design} A. To carry out the experiment, we mounted the metalens, and a detector onto a rotating stage. In order to image the DC component of the Fourier transform of an image, we placed a pinhole and a collimating lens in front of the meta-optic, and used an electronic display to image different scenes with the system. \autoref{fig:ms-design} B. shows the schematic of the optical setup.

\begin{figure}[t]
\centering
\includegraphics[width=\textwidth]{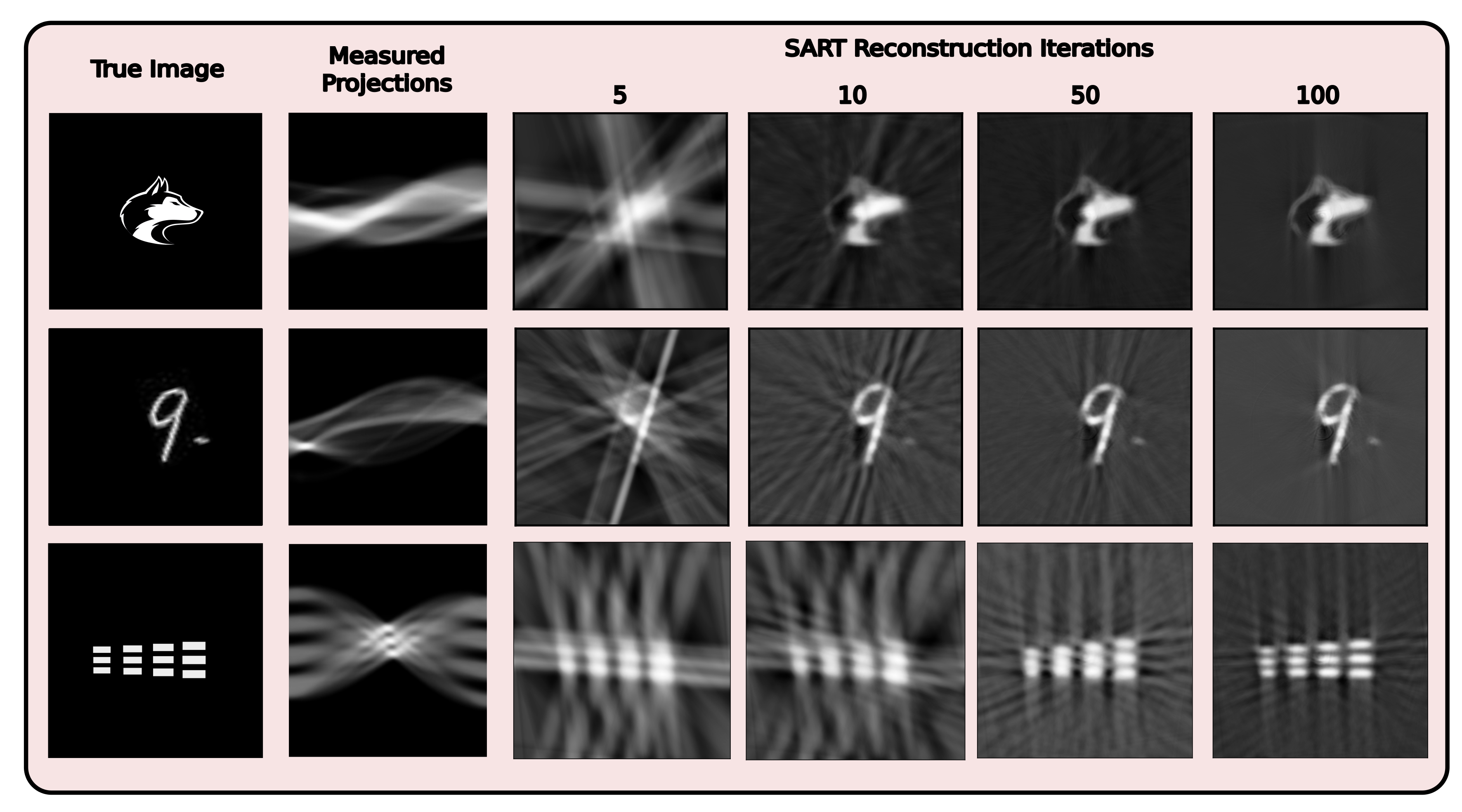}
\caption{Experimental results of metalens-based imaging using the radon transform. The left column is the true scene being imaged. Column second from the left are the raw measured sinograms collected experimentally. The x axis are the coordinates of the projection angle, and the y axis are the measured pixel values. The four rightmost columns are different steps in the reconstruction algorithm. Algorithm converges at about 100 iterations.}
\label{fig:results}
\end{figure}

The measurement was carried out by recording the line pixel data along different angular positions $\theta_i$ at the back focal plane of the cylindrical meta-optic. For measurement we used a regular $1936 \times 1216$ pixel detector, while only pixel values along the focal line were stored to emulate a line detector. In experiment, we stored $180$ different line projections corresponding to rotations between $0^o$ and $180^o$. To reconstruct the image we used the Simultaneous Algebraic Reconstruction Technique (SART) \cite{astra1,astra2,astra3}, which is a commonly used iterative reconstruction technique in medical imaging. \autoref{fig:results} shows 3 examples of measured Radon transforms and the reconstructed results using this technique, a picture of a husky, an MNIST number 9, and a horizontal resolution line target. We observe that the reconstruction converges after about 100 iterations, with no  significant improvemenet with subsequent iterations. Note that since we only measure $180 \times 1936$ pixels in the Radon domain, which corresponds to a $0.6\%$ compression ratio in the reconstructed $1936\times1936$ pixel image. Because the scenes we measured in this work are relatively dense (a single image essentially occupies most pixels) and lack many features, a simple average pooling filter which can be implemented by binning groups of nearby pixels achieves a similar imaging performance. Nevertheless, it is theoretically possible to achieve a better imaging performance on sparsely featured scenes (like a digit with a wide background) while capturing fewer pixels than in the original scene. For more details see \autoref{s3} and \autoref{fig:s3}.
\begin{figure}[t!]
\centering
\includegraphics[width=\textwidth]{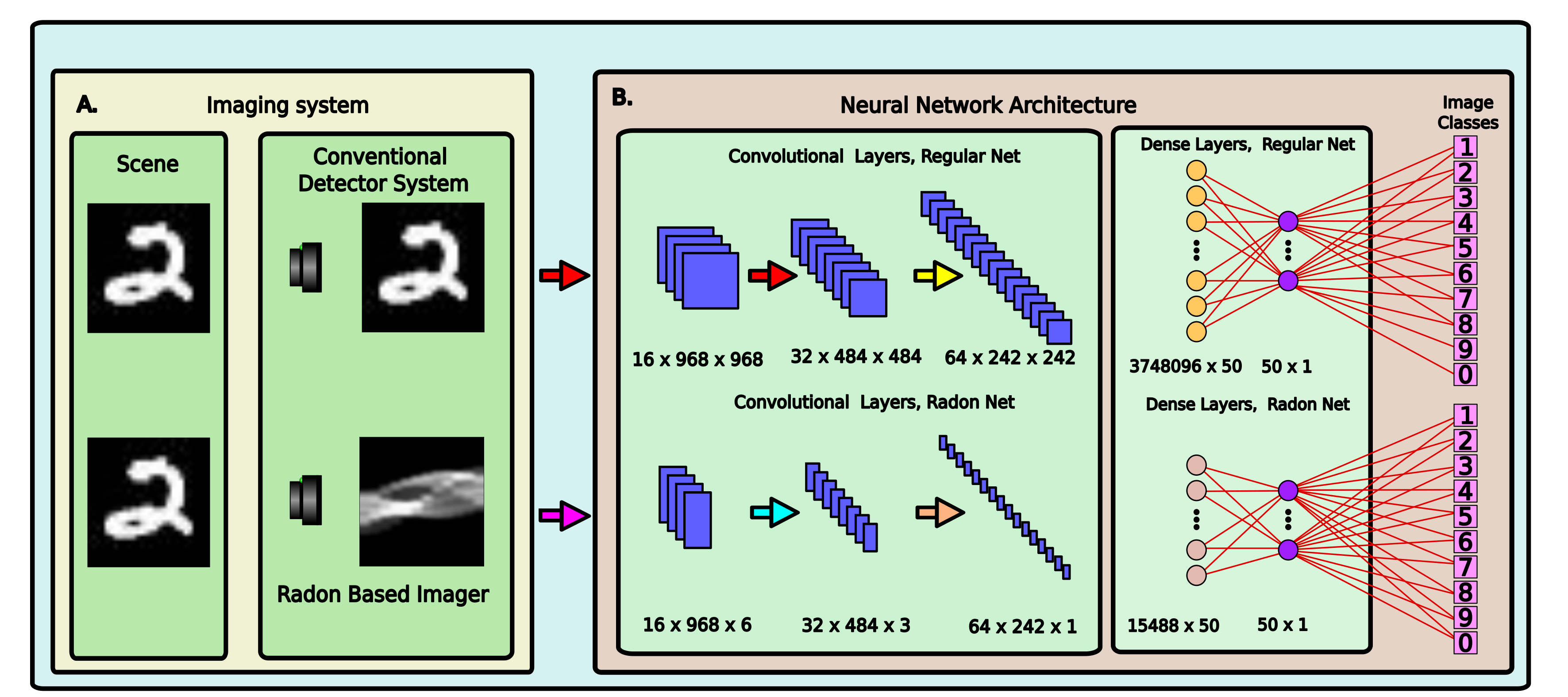}
\caption{\textbf{A.} The setup of the two imaging systems. \textbf{Top} shows a regular scene and detector system and an ideal output. \textbf{Bottom} shows the proposed Radon imaging system. \textbf{B.} Neural network architectures. \textbf{Top} conventional AlexNet based convolutions neural network with 3 convolutional layers, and 2 densely connected layers. \textbf{Bottom} Radon transform based neural network. The proposed layers are thinner due to the measured Radon transform containing fewer pixels in the angular dimension.}
\label{fig:nn}
\end{figure}
\section* {Neural network classification in Radon Space}

\begin{figure}[t!]
\centering
\includegraphics[width=\textwidth]{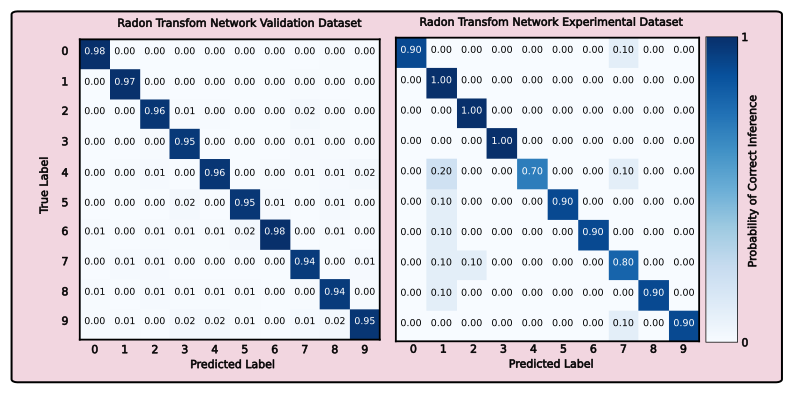}
\caption{Confusion matrices of the (left) validation set with radon transform computed numerically, and (right) experimentally measured radon transform.}
\label{fig:confusion}
\end{figure}

While reconstruction is often used to create an aesthetically pleasing image, for many machine vision problems, such as classification, we can directly utilize the captured data. By using a neural network-based back end, we can classify images in the radon space. For this task, we selected a common neural network architecture used in classification tasks called AlexNet \cite{AlexNet} consisting of convolutions, max pool layers, and dense layers. We used a modified version of AlexNet, with 3 convolutional layers and 2 dense layers. The neural network architecture is shown in \autoref{fig:nn}. To train the network we used the MNIST dataset. The MNIST data set consists of $28 \times 28$ pixel images of numbers 0-9. To train the network, we first padded the images by 14 pixels on each side to resemble the scale of the raw data captured in the experiment. Then, since our imaging setup uses a $1936 \times 1216$ pixel detector, we rescaled our images to fit this dimensionality. In order to correct for experimental rotation and misalignment, we performed random affine transforms of the real space images with rotations varying between $\pm50^o$ , translations between $10\%-30\%$ of the image width, and scaling between $80\%-120\%$ of the image size. To incorporate noise in our model, we added $5\%$ Gaussian noise to the raw images. Before feeding the images into the neural network, we radon transformed the images, keeping 23 angular projections in the sinogram evenly space between $0^o$ and $180^o$. After training, we achieved a $95\%$ classification accuracy on the validation set. To test the network with experimental data, we collected 100 experimental images of the MNIST data set (such as the number $9$ shown in \autoref{fig:results}) with the same angular spacing in the radon domain. Each classification class had 10 images. Classifying our experimental dataset, we achieved a classification accuracy of $90\%$.  \autoref{fig:confusion} shows the confusion matrices of the radon transform neural networks. We emphasize that, the network is not retrained with the experimental data, and training a network with only synthetic data generally yields much lower accuracy.

\newpage
\section*{Discussion}
In this paper, we presented and experimentally validated a meta-optical imaging system based on the Radon transform. The system is composed of a cylindrical metalens, a rotating stage, and a line detector which measured Radon projections of an image at different angles. The proposed system provides a key advancement in optical realizations of mathematical transforms with applications in accelerating image processing. We achieve imaging of a real scene, by capturing $0.6\%$ of the reconstructed image's pixels, effectively compressing the scene in the optical domain. The proposed system also offers a few advantages over conventional imaging systems. Firstly, line sensors with a large number of pixels in the order of $10^4$ are readily commercially available. This offers a unique advantage for the proposed imaging system over conventional 2D detector, which will be particularly well-suited for longer wavelengths, inclusing mid/ long wave infrared. By measuring projections of $N$ pixels per angle, an $N \times N$ pixel imaging system can be built without using $N \times N$ pixels, offering significant cost and power reduction, enabling $100-1000$ megapixel systems to be built efficiently.

The current setup is in fact a prototype and could be improved upon. It takes approximately 5-6 minutes to collect 180 Radon slices in order to reconstruct the full image. This is due to the long exposure time ($2$ seconds) needed to capture an image due to the inclusion of a pinhole. Furthermore, the rotating stage adds a moving part to the system which is less than ideal for realistic imaging setups.  Nevertheless, both of these issues can be avoided by manufacturing a two layer meta-lenslet array. One of the meta-lenslet array would have cylindrical phase profiles oriented at different angles. The other would have regular hyperboloid profiles effectively serving as the collimating lenses. Then we would use linear detectors at the focal plane of the system to collect the Radon transform in a single shot.

Another key aspect of this work shows the classification of images in a transformed domain. To the best of the authors' knowledge, most current works that use transformation optics as a front-end to a neural network, require both a simulated dataset, as well as experimental data that use those optics for calibration. This of course adds complexity to a system, and often even requires a digital calibration layer. In this work we demonstrate that experimental calibration is not strictly required if you carefully design a transformation optic based on mathematically known transforms that can easily be simulated on a computer.

\newpage
\section*{Fabrication Methods}
We used commercially available silicon on sapphire wafers with a film thickness of 500 nm from University Wafer. Wafer were diced into square chips with a side length of 2 cm. After thorough cleaning in an ultrasound bath of Acetone and subsequently IPA, we dry-cleaned the chips and used a further barrel etching step to remove organic residues. 
We then spin coated a positive resist (ZEP 520-A) onto the sample, with a subsequent baking step on a hot plate. To mitigate charging issues, a conductive polymer layer was further spun onto the substrate (DisCharge H2O). We then patterned the design into the resist using a 100 keV electron beam (JEOL JBX6300FS) at a dose of 275 muC cm-2. The conductive polymer layer was then removed in IPA and the chip was then developed using Amyl Acetate. Subsequently, we appplied another barrel etch step to remove remaining residues. After development, we deposited a layer of 50 nm alumina using an e-beam evaporator and subsequent lift off in NMP. After further cleaning, the silicon was etchined in a dry etcher using a mixture of C4F8/SF6.

\section*{Measurement Methods}
The cylindrical metalens was mounted one focal length in front of a sensor (ASI174MM) with a narrow band pass filter on a rotating mount. This imager was placed at the focal plane of a 4f system, consisting of an OLED display, a lens with 200 mm focal length, a pinhole with a diameter of 200 micron, and a second lens with a focal length of 50 mm. Thus the 4f system enabled a spatially coherent input for the radon imager, while also demagnifying the image by a factor of 4. All measurements were automated using custom written python scripts.

\bibliography{scibib}

\bibliographystyle{Science}



\clearpage

\end{document}



\baselineskip24pt


\maketitle



\begin{figure}[t]
\centering
\includegraphics[width=\textwidth]{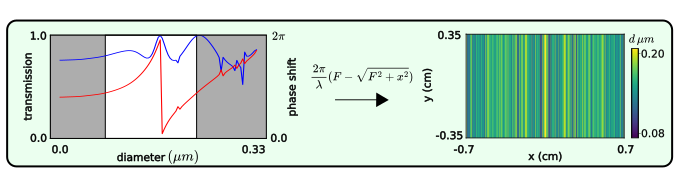}
\caption{Phase and amplitude response of Silicon square pillars. \textbf{b.} Designed metalens.}
\label{fig:s1}
\end{figure}

\newpage
\begin{figure}[t]
\centering
\includegraphics[width=\textwidth]{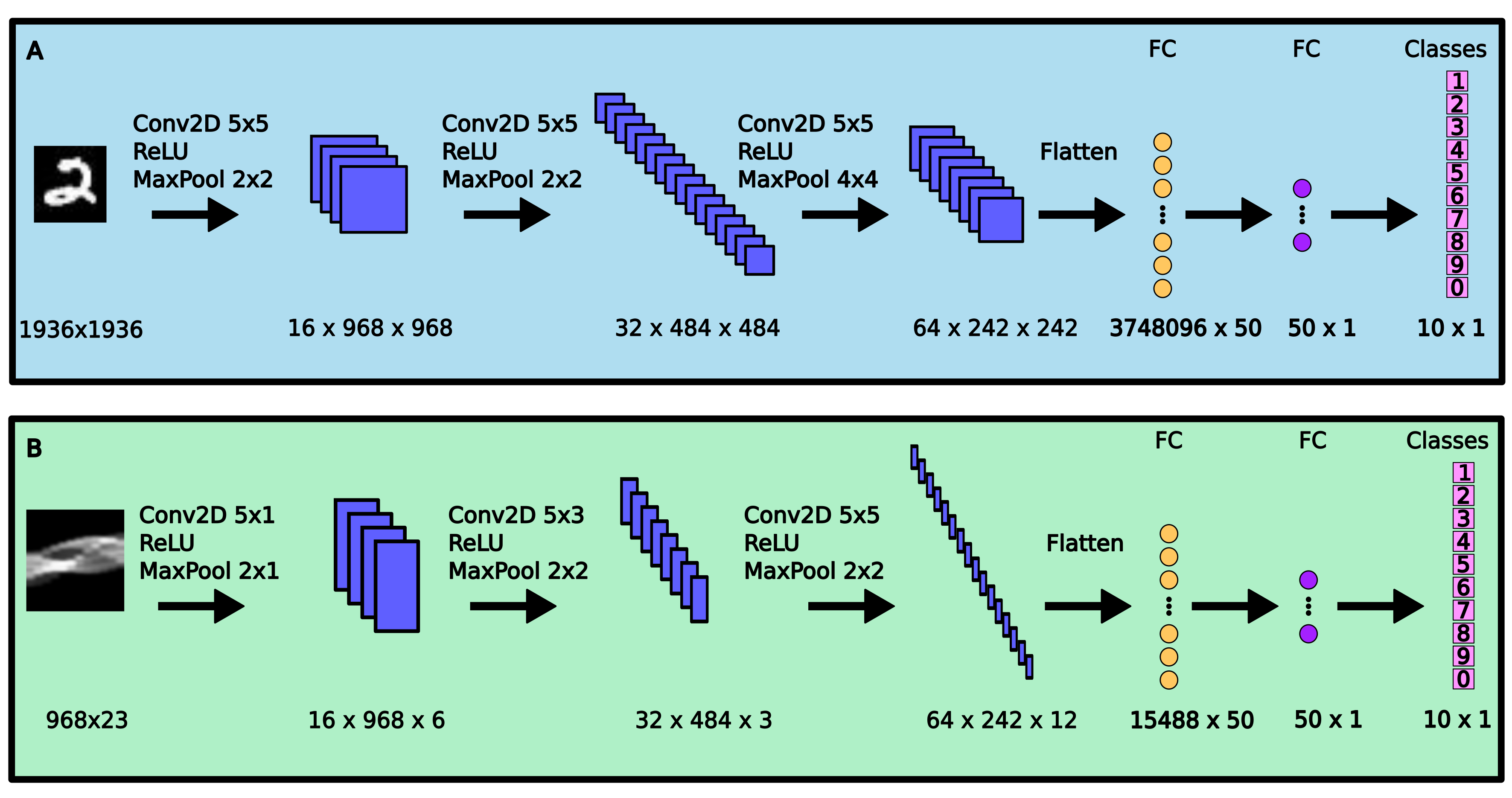}
\caption{\textbf{A.} Reference neural network for MNIST classification. \textbf{B.} Comparison neural network for classification in the Radon domain.}
\label{fig:s2}
\end{figure}
\newpage
\section{}
The scatterer library was simulated using the S4 package from Stanford. The material platform was chosen to be crystaline Silicon on Sapphire substrate. The height was set to $500nm$, and periodicity was set to $330nm$. The operating wavelength was chosen to be $780nm$. Due to manufacturing constraints, the pilar diameters in the design are constrained between $70$ and $202nm$.
\newpage
\section{}
The reference AlexNet neural network was designed to classify $1936 \times 1936$ pixel images of the MNIST dataset. It consists of 3 convolutional layers, and two fully connected layers as shown in \autoref{fig:s2} A. The neural network for classifying the Radon transformed images is a similar AlexNet architecture shown in \autoref{fig:s2} B. The key difference between the two architectures is since the Radon transformed data is thinner ($1936 \times 23$ pixels), the convolutional layers of the network are also thinner and rectangular instead of square. The datasets of both networks were prepared with random affine transformations to correct for experimental misalignment. The real space images were randomly rotated by $\pm^o$, translated by $10\% -30\%$, and scaled by $80\%-120\%$. To incorporate noise in our model, we added $5\%$ Gaussian noise to the raw images. The regular network was trained on real space data. The Radon network was trained by applying the Radon transform to the images, at $23$ evenly spaced angles.

\newpage
\section{}
\label{s3}
\begin{figure}[t]
\centering
\includegraphics[width=\textwidth]{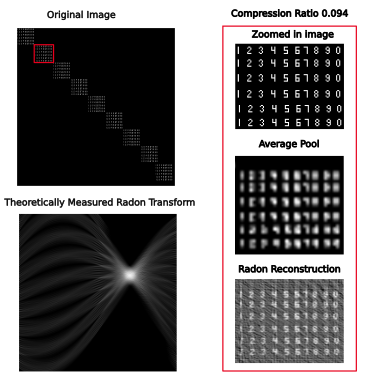}
\caption{\textbf{Left.} Top: $968 \times 968$ pixel image of densly packed numbers. Bottom: Theoretically computed Radon Transform with $91$ evenly spaced angles. \textbf{Right.} Top. Image zoomed in on one of the boxes of numbers. Middle: Result by average pooling with a compression ratio of $0.093$. Bottom: Image reconstructed via the Radon transform with the same compression Ratio}
\label{fig:s3}
\end{figure}
In this section we present a theoretical result for image compression via the Radon Transform compared to a baseline average pool method. This was done to see if it is theoretically possible to reconstruct images by measuring fewer pixels than in the original image and reconstruct the image of higher quality than a naive average approach. An average pool was chosen as a benchmark since optically, it is the simplest way of compressing an image by using several pixels as one and binning the result. The original image is an artificially generated $968\times968$ pixel image of numbers. Each number is drawn to have a maximum height of 7 pixels, and a width of 4 pixels. \autoref{fig:s3} shows the full $968\times968$ image and its zoomed in counterpart. To realistically compute the Radon transform optically, we took successive 1-D Fourier transforms of the image, and measured the DC component of it to simulate a cylindrical lens and measurement process. We measured $91$ different angles of the Radon transform. The theoretically computed Radon transform is shown in the bottom left of \autoref{fig:s3}. To reconstruct the compressed image, we took the inverse Radon transform shown in the bottom right of \autoref{fig:s3}. We computed the compression ratio as the number of pixels in the Radon transform, over the number of pixels in the reconstructed image $\frac{91 \times 968}{968^2} = 0.094$. The averaging window required to achieve a similar compression was $3\times 3$ achieves a similar compression ratio of $0.11$. As shown in \autoref{fig:s3}, the Radon reconstruction outperforms the average pool in this case, with the numbers being clearly resolved as compared to the average pool. We conclude it is theoretically possible to use this method to achieve image compression at the detector if we can achieve an accurate 1-D Fourier transform optically, which would require a cylindrical lens with a tight focal spot.